\documentclass{article}

\usepackage{arxiv}

\usepackage[utf8]{inputenc} % allow utf-8 input
\usepackage[T1]{fontenc}    % use 8-bit T1 fonts
\usepackage{hyperref}       % hyperlinks
\usepackage{url}            % simple URL typesetting
\usepackage{booktabs}       % professional-quality tables
\usepackage{pifont}% http://ctan.org/pkg/pifont
\usepackage[section]{placeins}
\usepackage{makecell}

\usepackage{rotating}

\usepackage{amsfonts}       % blackboard math symbols
\usepackage{nicefrac}       % compact symbols for 1/2, etc.
\usepackage{microtype}      % microtypography
\usepackage{lipsum}		% Can be removed after putting your text content
\usepackage{graphicx}
\usepackage{natbib}
\usepackage{doi}
\usepackage{url}

\usepackage{breakurl}

\title{Visual Knowledge Discovery with Artificial Intelligence: Challenges and Future Directions}

\author{
    \href{https://orcid.org/0000-0002-0995-9539}{\includegraphics[scale=0.06]{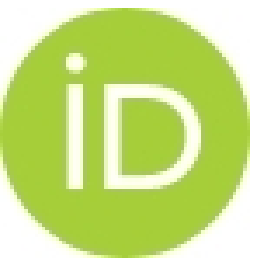}\hspace*{1mm}Boris Kovalerchuk}\\
	Central Washington University\\
	Ellensburg, WA, USA\\ 
	\texttt{boris.kovalerchuk@cwu.edu} \\
	\And
	\href{https://orcid.org/0000-0002-6015-3151}{\includegraphics[scale=0.06]{orcid.eps}\hspace*{1mm}R\u{a}zvan Andonie}\\
	Central Washington University\\
	Ellensburg, WA, USA, \\ 
	\& Transilvania University\\
    Brașov, Romania \\ 
	\texttt{razvan.andonie@cwu.edu} \\
	\AND
	\href{https://orcid.org/0000-0003-1600-0227}{\includegraphics[scale=0.06]{orcid.eps}\hspace*{1mm}Nuno Datia}\\
	ISEL-Instituto Superior de Engenharia de Lisboa\\
	\& NOVALINCS, Lisbon, Portugal \\
	\texttt{datia@isel.ipl.pt} \\
	\And
    \href{https://orcid.org/0000-0002-2907-2740}{\includegraphics[scale=0.06]{orcid.eps}\hspace*{1mm}Kawa Nazemi}\\
	Darmstadt University of Applied Sciences\\
	Darmstadt, Germany\\
	\texttt{kawa.nazemi@h-da.de} \\
	\And
	\href{https://orcid.org/0000-0002-9239-9420}{\includegraphics[scale=0.06]{orcid.eps}\hspace*{1mm}Ebad Banissi}\\
	London South Bank University\\
	London, UK\\
	\texttt{banisse@lsbu.ac.uk} \\
}

\renewcommand{\shorttitle}{Visual Knowledge Discovery with Artificial Intelligence}

\hypersetup{
pdftitle={A template for the arxiv style},
pdfsubject={q-bio.NC, q-bio.QM},
pdfauthor={David S.~Hippocampus, Elias D.~Striatum},
pdfkeywords={First keyword, Second keyword, More},
}

\begin{document}
\maketitle

\begin{abstract}
This volume is devoted to the emerging field of Integrated Visual Knowledge Discovery that combines advances in Artificial Intelligence/Machine Learning (AI/ML) and Visualization/Visual Analytics.  Chapters included are extended versions of the selected AI and Visual Analytics papers and related symposia at the recent International Information Visualization Conferences (IV2019 and IV2020).  AI/ML face a long-standing challenge of explaining models to humans. Models explanation is fundamentally human activity, not only an algorithmic one. In this chapter we aim to present challenges and future directions within the field of Visual Analytics, Visual Knowledge Discovery and AI/ML, and to discuss the role of visualization in visual AI/ML. In addition, we describe progress in emerging Full 2D ML, natural language processing, and AI/ML in multidimensional data aided by visual means.
\end{abstract}

% keywords can be removed
\keywords{Visual Knowledge Discovery \and Visualization of Artificial Intelligence/Machine Learning Models \and Visual Analytics}

\section{Introduction}
\label{sec:intro}

The human processes images 60,000 times faster than text \cite{Vogel1986}, and 90\% of information transmitted to the brain is visual \cite{Potter2014}. Moreover, Kelts reports that 35\% of our brain is devoted to vision \cite{Kelts2010}. A team of neuroscientists from MIT has found that the human brain can process entire images that the eye sees for as little as 13 ms\footnote{Anne Trafton, MIT News Office, January 16, 2014}. 

The definition given by the Merriam-Webster Dictionary for visualization is ``the formation of mental visual images, the act or process of interpreting visual terms, or of putting them into visible form''. Today, the actual content behind visualization and visual analytics is a collection of methods from multiple domains, including Artificial Intelligence/Machine Learning (AI/ML), rather than distinct methods. Within a broad definition, visualization is not a new phenomenon, being used in maps, scientific drawings, and data plots for hundreds of years \cite{Friendly2008}. Many visualization concepts are imported in computer visualization \cite{Tufte2001}. 

Data visualization is the process of translating raw data into images that allow us to gain insight into them. Common general types of data visualization are many ranging from simple charts to Timelines. A complete list and examples of interactive data visualizations can be found at\footnote{ 
https://www.tableau.com/learn/articles/interactive-map-and-data-visualization-examples}.  While this is true, the available tools are very limited in discovering multidimensional knowledge that is the core of ML. Therefore such definitions of information visualization inflate expectations beyond the actual capabilities. They ``oversell'' its current capabilities but rightly describe its future. 

The above definition of visualization is uncertain. It does not convey the significance of ``interpreting visual terms'' and identifying ``putting to the visual form''. An alternative definition tells us that visualization is (1) the act or an instance of visualizing or (2) (Psychology) a technique involving focusing on positive mental images to achieve a particular goal\footnote{https://www.thefreedictionary.com/visualization}. It is a bit more specific. The third definition is given in Wikipedia specifically for graphics: ``visualisation or visualization is any technique for creating images, diagrams, or animations to communicate a message'' really expresses the meaning that is common in computer science for it. It is clear from this definition that we \emph{visualize the message that already exists}. 

In AI/ML, the message of the main interest is a  \emph{pattern discovered} in the data and the \emph{prediction} for a new case based on this pattern. In other words, \emph{visualization is a visual representation of already existing data, information, and knowledge, but not a process of discovering new knowledge}. In contrast, the goal of Visual Analytics (VA) and Visual Knowledge Discovery (VKD) is broader; it includes \emph{discovering new messages -- knowledge using visual means beyond visualizing given input data, ML algorithm and already discovered patterns}. 

Thus, in the context of ML, we are interested in both Visualization of all existing AI/ML messages, and Visual Discovery of new AI/ML messages/knowledge, where visual means enhance AI/ML and AI/ML enhances visual means. In other words, we feel that it is time to expand the focus of visualization from visual message communication to message discovery with visualization means. It is hard to predict if the visualization term will cover both communication and discovery aspects or communication only. The terms of  visual analytics and visual knowledge discovery emerged to accommodate the discovery aspects. 

Tukey, who introduced Cooley the Fast Fourier Transform, suggested \cite{Tukey1977} that the idea of visualization helps us see what we have not noticed before. That is especially true when trying to identify relationships and find meaning in vast amounts of collected data: ``The greatest value of a picture is when it forces us to notice what we never expected to see.'' 

As we see, Tukey did not try to make a definition; he just noted that visualization allows us to see what we did not notice before in data. For example, we see numbers in the table and do not see any pattern, but visualization immediately shows a linear trend. Unfortunately, such impressive examples do not scale well to multidimensional data critical in AI/ML, as we already mentioned before. Thus, we should not exaggerate the abilities of the current tools and develop new tools which will be efficient for multidimensional pattern discovery in AI/ML. 

The term \emph{scientific visualization} refers to the process of representing scientific data. It provides an external aid to improve the interpretation of complex datasets, and to gain insights that may be overlooked by other methods (e.g., statistical methods).

Scientific visualization has evolved as a subset of computer graphics. The emphasis on visualization in computer graphics started in 1987 with the special issue of Computer Graphics, on Visualization in Scientific Computing. We quote from there: ``Scientists need an alternative to numbers. A technical reality today and a cognitive imperative tomorrow is the use of images. The ability of scientists to visualize complex computations and simulations is essential to ensure the integrity of analyses, to provoke insights and to communicate those insights with others.'' Very interesting, at the level of 1987, visualization was already considered as embracing both image understanding and image synthesis: Visualization is a method of computing. It transforms the symbolic into the geometric, enabling researchers to observe their simulations and computations. Visualization offers a method for seeing the unseen. It enriches the process of scientific discovery and fosters profound and unexpected insights.'' This is probably the most synthetic characteristic of scientific visualization: \emph{to see the unseen}. 

Since 1987, there have been several IEEE and ACM SIGGRAPH visualization conferences and workshops. Recent conferences on this topic include the International Conference on Information Visualization\footnote{ https://iv.csites.fct.unl.pt/au/, International Conference Information Visualization}. The Visualization Handbook \cite{Hansen2011} is a textbook that serves as a survey of the field of scientific visualization and computer graphics.

This chapter presents a vision of the future of VKD, including visual analytics. VKD can help humans understand how AI/ML algorithms learn and provide new knowledge discovery avenues. The chapter outlines the future of more traditional visual methods for developing and understanding multiple ML models. We summarize open problems and current research frontiers in visualization relevant to AI/ML.

\section{Visualization in ML}
\label{sec:vkd}

Visual analytics for ML has recently evolved as one of the latest areas in the field of visualization. A comprehensive review of progress and developments in visual analytics techniques for ML is available in \cite{Yuan2021}. The authors classified these techniques into three groups based on the ML pipeline: before, during, and after model building. 

According to Yaakov Bressler, Data Scientist at Open Broadway Data,  visualizing ML workflows most often only takes place at the final stage, but there are some situations where data visualization would take precedence during each step\footnote{https://www.quora.com/Why-is-data-visualization-essential-in-every-step-of-machine-learning}:

\begin{itemize}
\item \emph{In formulating a hypothesis}, exploratory data analysis helps contextualize problems.
\item \emph{Invalidating data integrity} by ensuring data is in the correct shape, we can create visualizations that demonstrate continuity, balance, or correct class.
\item \emph{Model selection}. Some ML scenarios will have widely different results depending on models' hyper-parameter selection/model architecture. Comparing these models' outputs (the loss functions) with visualization is a prevalent practice.
\item \emph{Testing model performance}. Once a model with good performance is identified, ML practitioners generally test this model to see how resistant it is to overfitting. Comparing performance is often done with data visualization.
\end{itemize}
\par
Today, visualization in ML appears in the following refined stages: 
\begin{enumerate}
\item New ML algorithm design stage, where visualization supports the \emph{design process of new ML algorithms} (e.g., designing a new ensemble algorithm from the existing algorithms using interactive visual programming with drag and drop).  

\item New model discovery by existing ML algorithm stage, where visualization supports the use of an algorithm to discover a model for the \emph{given data}:

\begin{itemize}
\item Visualization of \emph{input data} (this is similar to data visualization).
\item Visualization of the \emph{learning process} (e.g., how a decision tree algorithm learns the model by animation or other means).
\item Visualization of the \emph{results} (e.g., a learned model such as decision tree, SVM, CNN and saliency map). 
\item Visualization of \emph{learning process refinement} (e.g., pruning a decision tree model).  
\end{itemize}
\end{enumerate}

Stages 1 and 2 assume the \emph{current paradigm} that the ML algorithm operates in the n-dimensional data space, not in the 2-D or 3-D visualization space. It is apparent why 2D or 3D visualization space was not used in Stages 1 and 2. Traditionally visualization space is lossy, and it does not preserve all n-D information \cite{Kovalerchuk2018}. 

The situation has changed with building a 2-D/3-D visualization space that preserves all n-D information \cite{Kovalerchuk2018}. This creates a \emph{new paradigm} of visual knowledge discovery in this 2-D/3-D visualization space that we call visual space for short. This paradigm is discussed in Sections 3.4 and 4 below, and in \cite{Kovalerchuk2021}.

There are very few monographs on the visualization of ML algorithms \cite{Simoff2008}. The recent monographs focus primarily on applications using different languages: Python \&  Mathematica \cite{Awange2020}, R \cite{Wiley2019}, Julia \cite{Salceanu2018}, and Scala \cite{Manivannan2015}. Dedicated software tools are also available\footnote{https://neptune.ai/blog/the-best-tools-for-machine-learning-model-visualization}. For instance, MLDemos\footnote{https://basilio.dev/} creates an open-source visualization tool for machine learning algorithms in order to understand how several algorithms function; how their parameters affect and modify the results in problems of classification, regression, clustering, dimensionality reduction, dynamical systems and reward maximization. 

In\footnote{https://towardsdatascience.com/machine-learning-visualization-fcc39a1e376a}, it was pointed out that:
\begin{quote}
One of the main limitations of plotting decision boundaries is that they can only be easily visualized in two or three dimensions. Due to these limitations, it might be necessary \emph{to reduce the dimensionality} of our input features (using some form of feature extraction techniques) before plotting the decision boundary.
\end{quote}

We want to emphasize the part that we put in italic above on reducing the data dimensionality. It represents typical mainstream practice, not an emerging opportunity. This work did not mention the negative effect of Dimension Reduction, which is generally lossy, as Johnson-Lindenstrauss Lemma \cite{Kovalerchuk2018} shows. Therefore, 2-D decision boundaries likely will distort the n-D boundary that was built by the ML methods. The above citation is from a paper published in October 2020, but demonstrated before in \cite{Kovalerchuk2012, Kovalerchuk2018}, that a lossy Dimension Reduction is not necessary to visualize the n-D decision boundary fully. The concept is further expanded in this volume, in \emph{Non-linear Visual Knowledge Discovery with Elliptic Paired Coordinates} by McDonald and Kovalerchuk, and in \emph{Self-service Data Classification Using Interactive Visualization and Interpretable Machine Learning} by Wagle and Kovalerchuk. 

These works had shown that n-D decision boundaries could be represented in 2-D/3-D fully without any information loss of n-D. Therefore, one of the goals of this volume is to attract attention to these new methods that enhance ML capabilities for applications.

These lossy situations only reflect the mainstream practice, not an emerging opportunity to use visual methods as a core of the ML model development that we call VKD.    

Visualization of knowledge extraction is a topic with a long history in scientific visualization (see \cite{Bonneau2006}). For instance, a representative paper from that volume extracts the isosurfaces in 3-D data.   
  
However, the topic of visualization of ML outcomes is new, ``hot'', and growing. A search on IEEE Xplore for ``visualization'' + ``machine learning'' in the title results in 43 entries. Of these, over 85\% are from the last five years. In contrast, ``visualization'' + ``data'' has 2,200 entries. Of these, less than 40\% are from the last five years. ``Artificial intelligence'' and ``visualization'' have only nine entries.

Recently, a new journal was introduced: Machine Learning and Knowledge Extraction\footnote{https://www.mdpi.com/journal/make} (vol 1 in 2019). This journal fosters an integrated machine learning approach, supporting the whole ML and knowledge extraction and discovery pipeline from data pre-processing to visualizing the results. ``Visual Informatics'', another new journal published by Elsevier, is dedicated to the visual data acquisition, analysis, synthesis, perception, enhancement and applications\footnote{https://www.journals.elsevier.com/visual-informatics}. 

Looking more closely at the actual visualizations used with ML, we observe that many of them are limited to heatmaps (for saliency maps) and bar charts (for feature importance). Therefore, we can say that the synthesis of ML and visualization is still in its infancy as an interdisciplinary domain, with an expectation of broader integration with a range of real-world applications in the future. 

The importance of visual methods in ML has grown recently due to their perceptual advantages over other alternatives for model discovery, development, verification, and interpretation. In ML-model development and verification, visual methods are beneficial to avoid both overgeneralization and overfitting. In ML model interpretation, the human component is a significant part of this process. It is fundamentally human activity, not a mechanical activity. Visuals can naturally support efficient ML explanation. The other process in this area requires overcoming several limitations, such as human understanding of complex multidimensional data and models without ``downgrading'' it to human perceptual and cognitive limits. Existing methods often lead to the loss of interpretable information, occlusion, clutter and result in quasi-explanations \cite{Kovalerchuk2009}.

A list of six challenges and potential research directions in ML visualization was suggested in \cite{Yuan2021}: improving data quality for weakly supervised learning and explainable feature engineering before model building, online training diagnosis and intelligent model refinement during model building, and multimodal data understanding and concept drift analysis after model building.

\section{Visual Analytics, Visual Knowledge Discovery, and AI/ML}
\label{sec:visual-analytics}

Artificial intelligence, machine learning, and visualization are the technological drive for visual knowledge discovery within the application domain.
%that the tasks are manual intensive for users. 
This section presents different aspects of the current and future of this mutual enrichment.

\subsection{What is Visual Analytics?}
\label{subsec:VA}
Visual Analytics systems combine AI/ML automated analysis techniques with interactive data visualization to promote analytical reasoning. In this case, the focus is on interactive data analytics using machine learning rather than visualizing machine learning.

Thomas and Cook proposed an early definition of Visual Analytics as ``the science of analytical reasoning facilitated by interactive visual interfaces'' \cite[p. 4]{TC05}. They emphasized the “overwhelming amounts of disparate, conflicting, and dynamic information” \cite[p. 2]{TC05} in particular for security-related analysis tasks. Visual Analytics focused thereby mainly on \emph{``detecting the expected and discovering the unexpected}” \cite[p. 4]{TC05} from massive and ambiguous data. Thomas and Cook outlined that the main areas of the interdisciplinary field of Visual Analytics are (1) analytical reasoning, (2) visual representation and interaction techniques, (3) data representation and transformation, and (4) production, presentation, and dissemination \cite{TC05}. Thereby analytical reasoning should enable users to gain \emph{insights} that support \emph{“assessment, planning, and decision making”}. 
The visual representation and subsequently the interaction techniques should enable users to “see, explore, and understand large amounts of information at once”. The data transformation process converts dynamic and heterogeneous data in a way supported by visualizations, production, presentation, and dissemination. It allows to communicate the results of the analysis process to a broad audience \cite{TC05}. Their definition and the related process should be outlined to focus on information rather than on raw and unstructured data. This indicates that the raw and unstructured data from heterogeneous resources are already synthesized and processed to get information. In this context, information is synthesized and processed data that can easily be visualized. The transformation process synthesizes data from different sources and different types into a unified “data representation” that can be interpreted as information \cite[p.11]{TC05}.
Visual Analytics gained throughout the years a series of revised and more specific definitions. Keim et al. commented that defining such an interdisciplinary field is not easy \cite{KKEM10} and proposed a different definition:

\begin{quote}
    Visual analytics combines automated analysis techniques with \emph{interactive visualizations} for an effective \emph{understanding, reasoning and decision making} on the basis of very large and complex data-sets. \cite[p. 7]{KKEM10}
\end{quote}

They suggested the combined use of \emph{automated analysis methods} and interactive visualization, particularly for understanding, reasoning, and decision making. The \emph{automated analysis} in this context relied on \emph{data mining approaches} \cite[pp. 41]{KKEM10} based on the work of Bertini and Lalanne \cite{BL10}. They differentiate between the data mining and information visualization process. The data mining process incorporates the steps from data to computational model by transforming data into a computational model, which allows the interpretation and verification of data and generating hypotheses that lead to knowledge. This process has no feedback loops. In contrast, information visualization incorporates the steps of mapping data to a visual model, which allows pattern extraction for generating hypotheses that leads to knowledge and has a feedback loop to all previous steps. This process is based on the initial work of Card et al., who proposed the information visualization reference model \cite{CMS99} with the steps of data transformation, visual mappings, and view transformation, including a feedback loop to all steps.  

Keim et al. proposed a Visual Analytics model, based on the introduced two processes of data mining and information visualization \cite{KKEM10}. This combined model starts with data and spreads out on two paths: the path to visualization uses mapping, and the path to models uses data mining \cite[p. 10]{KKEM10}. The main difference is that they included a direct combining of visualizations and models. The path can first go through the data mining path to the computational model and be visualized. Vice versa, a visual mapping can be used for model building \cite{KKEM10}.
The entire Visual Analytics process provides an interactive process to make use of both the interactive visual representations and data modeling approaches for acquiring knowledge and insights \cite{KKEM10}. The role of humans and the possibilities to interact in the stages of the Visual Analytics process remains as they are proposed in the reference model for visualization \cite{CMS99}. The main difference is the interactively combined techniques for visualizing and mining data.
In the past years, the limitation of Visual Analytics was lifted by different approaches, particularly machine learning and artificial intelligence. Recent works integrate decision trees and rule-based classifiers \cite{SMSS*21}, time series analysis \cite{MBSW*21}, topic modeling and word embedding \cite{EKCK*20}, statistical methods, machine learning \cite{CWM15},\cite{NB19}, and neural networks \cite{PHGL*18}.

The emerging coupling of artificial intelligence and machine learning approaches in Visual Analytics leads to discover new patterns, new knowledge and enables humans to facilitate in-depth analysis. With such a direct combining of artificial intelligence, machine learning, and interactive visualization methods, Visual Analytics is more than just the “science of analytics reasoning facilitated by interactive visual interfaces”, as proposed by Thomas and Cook \cite{TC05}. It has \emph{changed its problem-solving context} dramatically since the processing of vast amounts of raw data is an integral part of Visual Analytics. Considering that interactive visual interfaces can not lead to analytical reasoning, we define Visual Analytics as follow: 
\begin{quote}
    Visual Analytics is the science of analytical reasoning facilitated by the direct coupling of learning models and information visualization. 
\end{quote}

Thereby “learning models” consist of any learning method, e.g., unsupervised, semi-supervised, and supervised learning. Information Visualization is, per definition, interactive, as already proposed by Card et al. \cite{CMS99}. Visual Analytics should be seen as a discipline that incorporates both humans and computers, whereas the advanced automatic processing of any data is prominent.

\subsection{Human Interaction}
\label{subsec:HI}

\begin{figure}[htb]
\centering
\includegraphics[width=1\textwidth]{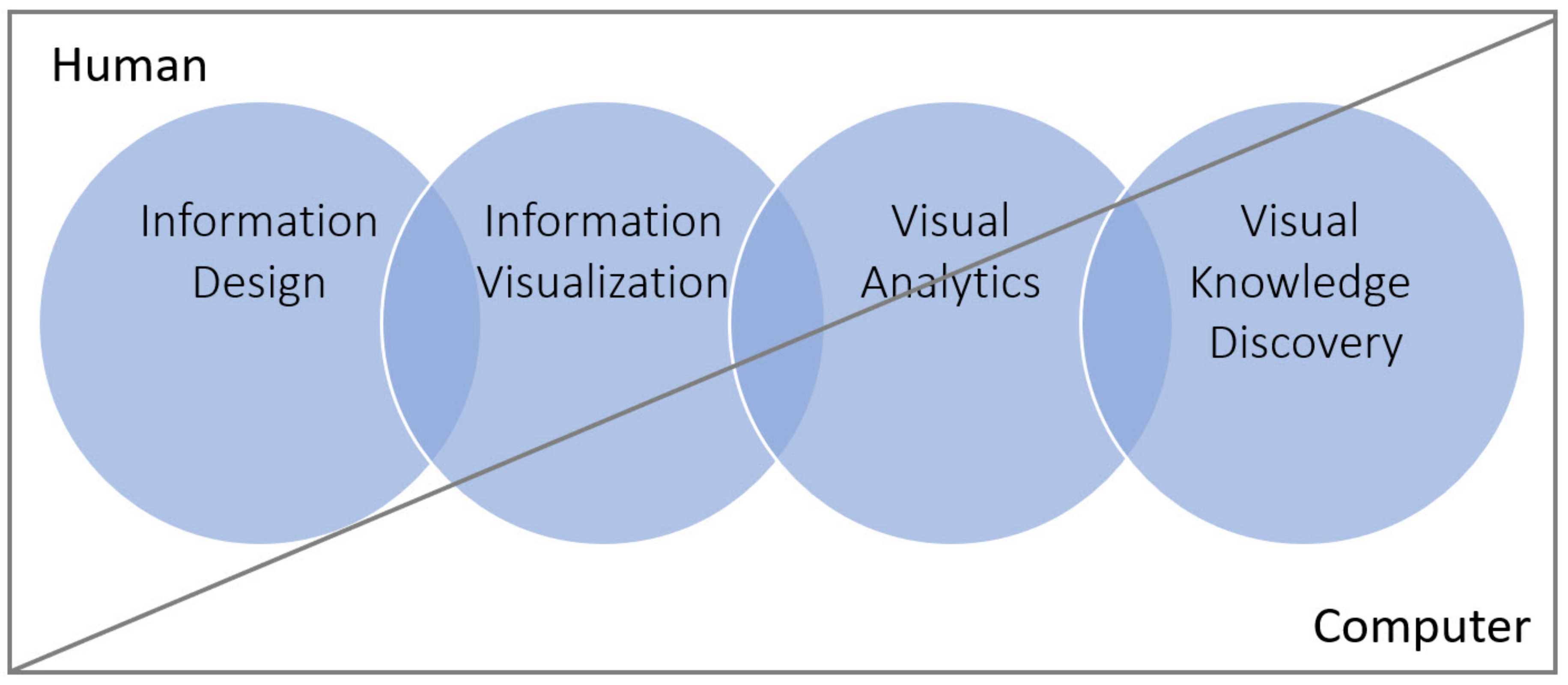}
\caption{The humans‘s and computer’s roles in visualization and knowledge discovery disciplines.}
\label{fig:va}
\end{figure}

Figure \ref{fig:va} illustrates the computer’s and human’s roles based on our previous work \cite{KNRB12}. Below we discuss the change of the human roles from Information design to visual knowledge discovery.   

%__________________________________________________________________
%_________________________________________________________________  

\emph{Information Design} (ID) or typically non-interactive ``\emph{Info-graphics}'' are static, non-interactive visual representations of mostly abstract data. The computer has no active role here since the static graphics represent information graphically without interaction capabilities and without pre-processing of data. Info-graphics may also occur in non-computational media. Information Visualization is the interactive visualization of abstract data. The computer pre-processes the data, and the interactivity of the visual interfaces is also enabled through the computer. However, \emph{Information Visualization commonly does not integrate any learning methods}. Visual Analytics commonly incorporates different pre-processing and learning methods. It couples machine learning methods directly with interactive visualizations and enables them to choose and parametrize the learning methods in the best cases. VKD goes one \emph{step beyond} by \emph{deeper coupling advanced learning methods} as described in Section \ref{subsec:vkd}. 

Commonly humans' interaction with an interactive system refers to Human-Computer Interaction (HCI). The interaction modalities of HCI vary enormously from simple interactive systems like text-editors to interactions with advanced automated computer systems like ML-based systems.  While many types of interactions are common for both, fundamental differences exist. Advanced visual learning systems based on machine learning/artificial intelligence methods require much more human involvement and more complex interaction modalities and goes far beyond the influential ``information-seeking mantra'' proposed by Shneiderman \cite{Shn96}: Overview First, then Zoom and Filter, and followed by Details on Demand.

Below we discuss the types of human interaction with automated systems from both human and automated system perspectives in Information Design (ID), Information Visualization (IV), Visual Analytics (VA) and Visual Knowledge Discovery (VKD). For the human, the automated system is an \emph{assistant}, while for \emph{the automated system}, the human is one of the \emph{sources} of information.
The\emph{ human perspective} of types of interactions with an automated visual system can be categorized as follow:

\begin{enumerate}

\item In Information Design, a human does not interact with the visual system. The human takes the role of the information \emph{consumer}.

\item In \emph{Information Visualization}, a human decides which visualization to use and interacts with the visualizations, e.g., through panning, brushing, zooming, and deriving conclusions. The human takes the role of information \emph{consumer}, while simple interactions do not change the information behavior.

\item In \emph{Visual Analytics (VA)}, a human first defines the problem to the level sufficient to use learning methods and other analytical tools, likely in several iterations. Next, the human plans the work of such automated systems, delegates some tasks to them, and analyzes the results. This is the mixed role of a direct \emph{creator}, a \emph{planner}, and an \emph{analyzer}. 

\item In \emph{Visual Knowledge Discovery (VKD)}, as a field of VA and AI, the role of the human is the same as a mixture of a \emph{creator}, a \emph{planner}, and an \emph{analyzer}, but with important specifics of guiding specialized advanced automated \emph{Visual Knowledge Discovery} ML-tools. See examples in Sections \ref{sec:Full2D} and \ref{sec:NLP},  where a human interactively discovers and analyzes patterns in the lossless visual space provided by the General Line Coordinates. 
\end{enumerate}

In (3) and (4), more time is devoted to \emph{interactive planning }and \emph{managing} the work of the advanced automated ML systems. In making such interactions efficient, \emph{new methodologies, approaches, interactive protocols, and tools} are needed.   
From \emph{AI/ML and systems perspective}, the types of interactions with humans are defined by the human's need as a source of data, information, and knowledge for such systems. Typically, the primary ML data (training, validation, and testing data) are generated in advance without entering them interactively. However, the ML system needs other information from humans. \emph{Explaining} the machine learning model and its prediction in terms of \emph{domain knowledge} nowadays needs a domain expert as a source of this knowledge, especially if such knowledge is implicit/tacit. Improving \emph{the accuracy} of the model prediction also still needs a domain expert as a source of this knowledge that includes tips to modify attributes, remove irrelevant attributes, search for specific patterns. Both explanation and improving accuracy are tasks with multiple uncertainties. Thus, human interaction is needed when we have \emph{uncertain} tasks that are not ready for automation. 

\subsection{Future Challenges of Visual Analytics}
As the science of analytical reasoning, Visual Analytics combines two main research areas that lead to a variety of future challenges: On the one hand, the computational learning models, and on the other hand, interactive visualization that involves humans in the entire transformation process. The following challenges need to be addressed: 

\ \\
\noindent \textbf{Adaptive Visual Analytics}: Considering that humans perform the analytical reasoning process through complex interactive visual representations, it is necessary to consider the cognitive, mental, and perceptive capabilities of humans. Besides considering humans' abilities, their interests, tasks to be solved, and the data (content) are important to consider. Adaptive Visual Analytics uses machine learning methods to adapt the visual interface, the data, and the interaction concepts to the demands of a specific user or user group \cite{Na16}. It reduces the complexity of analytical reasoning tasks and leads to more efficient and effective problem-solving. Although several approaches and even implemented systems exist \cite{Na18}, the increasingly complex and visual representations require further research to meet the demands of a specific user or a user group.

\ \\
\noindent \textbf{Visual parametrization and model adjustment}: The direct and deep coupling of learning methods and interactive visualizations enables visual parametrization and learning model adjustment to improve the learning results and to reduce overfitting. Visual Analytics should integrate visual interaction on model-level to allow refinements and model adjustments of the integrated learning method. The adjustment and refinement of the underlying learning method would lead to better results and a better understanding of the integrated methods.

\ \\
\noindent \textbf{Multi-model integration}: Visual Analytics allows integrating a variety of interactive visualization in a juxtaposed and superimposed way. Besides integrating interactive visualizations, Visual Analytics allows integrating more than one learning method, e.g., for data processing, forecasting, or clustering. While single-model Visual Analytics systems provide commonly appropriate analytical reasoning, the problem context and task-solving are strictly limited. Multi-model Visual Analytics should integrate a variety of learning models that allow solving analytical tasks with different models. Ideally, the choice of the underlying model can be performed by users through the interactive visual interface. 

\ \\
\noindent \textbf{Assisted Visual Analytics}: Analytical reasoning tasks may be difficult to solve even with appropriate interactive visualizations and learning methods. Visual Analytics systems could support users through guidance and assistance. Although assisted Visual Analytics seems to be similar to Adaptive Visual Analytics, the main difference is the fact that users get guidance but no automatic adaptation. The complexity of Visual Analytics increases with the number of visual layouts and learning models. Users should get recommendations based on tasks to be solved, data, integrated learning methods, and particularly the capabilities of the integrated visual layouts and learning methods.

\ \\
\noindent \textbf{Progressive Visual Analytics}: Coupling learning methods and interactive visualizations support the analytical reasoning process in Visual Analytics. However, the reasoning is performed by users. With progressive Visual Analytics, users get intermediate results of the underlying learning methods or any integrated models \cite{ASSS18}. The intermediate results are particularly of great interest if supervised learning methods are integrated into a Visual Analytics system. The computational processes are getting more transparent through execution feedback and control \cite{MPGS*14}. Furthermore, the intermediate results lead particularly in multi-model Visual Analytics to early decisions \cite{FPDS14}, parameter refinements, and model optimization through interactive visualizations. Progressive Visual Analytics should be applied in multi-model visual analytics systems that allow refinements and parametrization of the underlying learning methods.

\ \\
\noindent \textbf{Explainable AI through Visual Analytics}: Explainable AI has been gained a lot of popularity in research. It provides information about how and why the inputs led to certain outputs \cite{JKP19} and increase transparency, result tracing, and model improvement \cite{PORO20}. Through the direct combining of learning methods with interactive visualizations, Visual Analytics is predestined for explainable artificial intelligence \cite{Kovalerchuk2009}. Particularly multi-model Visual Analytics systems and those that use supervised methods can use the interactive visualizations of Visual Analytics to explain the underlying learning methods through visual interfaces.

\subsection{Visual Knowledge Discovery as an integral part of Visual Analytics and Machine Learning}
\label{subsec:vkd}
Visual Knowledge Discovery (VKD) is a part of a wider field of Visual Analytics (VA). 
Respectively the VA includes more tasks and goals that are in the realm of VKD. VA includes any task where we use a combination of analytical and visual tools to solve it. VA tasks can be decision making, prediction, preliminary data analysis, post prediction analysis, post decision making, more exact problem formulation for very uncertain situations are needed. It is a vast field, which is a source of VA's strengths and weaknesses. It is challenging to develop such a diverse field deeply enough in all its parts simultaneously because VA borrows methods from very diverse fields that are also not equally well developed and often with competing methodologies.  
While VKD also can be given the same far-reaching meaning as it is done for VA, we prefer to keep a scope of VKD narrower, keeping its meaning closely associated with the concept of the knowledge discovery in databases (KDD) used in the machine learning/data mining (ML/DM) community for predictive supervised and unsupervised ML/DM modeling. Therefore, we view VKD as an intersection of VA and ML/DM domains, where we view ML/DM as a part of the broader AI domain.
If we do not give VKD this narrower meaning, we can view it much broader, practically equal to VA. For instance, discovering or formulating the decision strategy can also be considered a part of the knowledge discovery. At this time, we prefer a narrower meaning of VKD as a field of visual predictive modeling integrated with supervised and unsupervised ML/DM methodologies.     
To clarify the differences between the domains, consider an example of stock market trading. In this task, we need to make an investment decision by picking up a set of stocks. 

First, we need to collect the data, predict the behavior of different stocks on the market, and finally make an actual investment decision. Different investment strategies lead to different investment decisions. Those investment strategies are partially based on the prediction and its quality. The low quality of the prediction leads to more conservative strategies than a high-quality prediction. In detail, such an example is considered in \cite{Wilinski2017} for USD and EURO trading using VKD as a part of the whole process.  
Hence, we focus on visual knowledge discovery with visual predictive modeling because the models are not accurate enough and not well-explained. The user cannot trust them and hardly will have solid, well-grounded further steps of VA such as decision making.
Traditional visualization methods, which convert n-D data to 2-D data, are lossy, not preserving all multidimensional information \cite{Kovalerchuk2018, Kovalerchuk2009}. In contrast, the representation of n-D data using General Line Coordinates (GLC) is lossless \cite{Kovalerchuk2018}. This visual representation of n-D data opened the opportunity for full multidimensional machine learning in two dimensions without loss of information (Full 2-D ML).

In simple situations, it allows visually discovering the pattern by observing these n-D data visualized in GLC. In more complex situations, which are common in ML, it allows discovering patterns in 2-D representations using new 2-D ML methods. We envision that it will grow into a whole new field of full 2-D machine learning. It requires developing a whole new class of 2-D ML methods. The firsts set of such methods have been developed and are presented in this volume with different types of General Line Coordinates (Parallel Coordinates, Radial Coordinates, Shifted Paired Coordinates, Elliptic Paired Coordinates, In-Line Coordinates, CPC-R and GLC-L) \cite{Kovalerchuk2018, Kovalerchuk2021, Dovhalets2018, Kovalerchuk2020a, Kovalerchuk2018Springer, McDonald2020, WagleK20}. The studies in this realm can be traced to Parallel Coordinates \cite{Inselberg1998, Inselberg2009}. 

Often 2-D studies in ML cover only simple 2-D examples to illustrate ML algorithms visually. Next, Visual Analytics studies have been very active in exploring parallel coordinates for tasks related to clustering \cite{Hohman19}, but much less for supervised learning, which needs to be developed further. The studies in this area include \cite{Estivill-Castro2020, Gary2017, Xu2007}. We believe that it is time to consolidate all such studies within a general concept of a full 2D ML methodology. Traditionally 2D studies in machine learning were considered only an auxiliary exploratory data/model visualization with loss of n-D information mostly afterward or before the actual machine learning. It was assumed that in 2-D, we are losing n-D information, and we need complete n-dimensional analysis in n-D space to construct ML models. The full 2-D ML methodology shows that it is not necessary. 
It expands visual discovery by human-aided ML methods to the full scope of machine learning methods to visualize full patterns analysis and with 2D interaction.

\section{Full 2D Machine Learning with Visual Means}
\label{sec:Full2D}

As of today, clustering tasks dominate in visual analytics: ``Clustering is one of the most popular algorithms that have been integrated within visual analytics applications'' \cite{Endert2017}. In the IEEE published 2019 survey of clustering and classification for time series in visual analytics \cite{Ali2019}, the summary table shows 79 papers on clustering and only seven papers on classification (decision trees, SVM, neural networks and others) from 2007 to 2018, i.e., over 10 times dominance of clustering. The earlier review (2017) on state of the art in integrating ML into visual analytics \cite{Endert2017} shows 26 papers on clustering, 13 on classification, and nine on regression in the summary table. Together only 21 papers are on supervised learning vs. 26 papers on clustering. All 21 papers on supervised learning (classification and regression) focus on modifying parameters, computation domain, and analytical expectation, i.e. pre and post modeling, not actual model construction tools, which are traditional \emph{existing} ML algorithms in these papers. 

Why is it important to change the focus in \emph{visualization} for ML from unsupervised learning (clustering, unsupervised dimension reduction) to supervised learning (classification, regression, supervised dimension reduction)?   It is not a move from one type of ML task to another equally important task, but it is the change of the research goal. Major impressive current achievements of ML are in supervised learning not in clustering. The fundamental difference between supervised and unsupervised learning is that, in supervised learning, we have a basis for judging the quality of our solution (i.e., how good or bad the solution is). In contrast, unsupervised learning is considered ill-posed \cite{Endert2017}. Therefore, with supervised learning we can progress more efficiently and solve important predictive problems, like medical diagnostics. In clustering, we rarely, if ever, solve such “final” problems.  

On the other side, the role of clustering is growing in ML. Many funding agencies now support extensive activities to create large databases of  labeled cases, especially in medicine and health care. However, it is a long and expensive process. Clustering is considered as a promising and less expensive approach to assist in solving this problem.  For supervised learning, clustering is an important auxiliary supporting task. It is not a predictive task per se, while it has multiple benefits for improving supervised learning. In between, we have semi-supervised learning tasks with some cases unlabeled.

In summary, clustering is not the primary ``final'' task; it does not predict classes. It only helps to predict, while supervised learning predicts. Therefore, supervised learning is the core of machine learning and focusing on it is well justified for visual analytics and visual knowledge discovery. While clustering plays only a supporting role in supervised learning predictive tasks, it is beneficial in many other tasks like pointing to the outliers that may or may not be related to predictive tasks. 

The third review from 2018 on visual analytics in Deep Learning (DL) \cite{Hohman19} outlines the next frontiers for deep supervised learning in using visualization during and after training DL models, again assuming existing DL models. A similar idea is presented in the 2020 review \cite{Eisler2020} with visualization for existing ML models. The most recent review from 2021 on visual methods in ML model interpretation \cite{Kovalerchuk2009} pointed out that the visualization methods used in DL are very limited. Typically they are  heatmaps for activations and bar charts for   feature importance.

The goal of full 2D machine learning is different. It is developing \emph{new ML models} based on the lossless 2D visual representation (visualization) of n-D data. In this methodology, visuals move \emph{from supporting tools to core knowledge discovery tools}. 

Below we present a representative but not a complete list of open problems of full 2D machine learning that we envision as a new research frontier. 

\begin{description}

\item[\textbf{Problem 1}] Developing \emph{new lossless visualizations} of multidimensional data. Several types of General Line Coordinates \cite{Kovalerchuk2018} are developed, explored, and applied at different levels (parallel coordinates, radial coordinates, shifted paired coordinates, collocated pairs coordinates, in-line coordinates, GLC-L) \cite{Kovalerchuk2018, Kovalerchuk2009, Kovalerchuk2021, Dovhalets2018, Kovalerchuk2020a, Kovalerchuk2018Springer, McDonald2020,  WagleK20}.  Many other GLCs are just defined in \cite{Kovalerchuk2018}.  This includes n-gon coordinates and other not paired GLCs,  which  recently have been implemented as npGLC-Vis  software library in \cite{Luque2021}.  

Developing new lossless visualizations of multidimensional data should not be limited by GLCs. Other lossless visual representations of multidimensional data are possible, and some already exist \cite{Kovalerchuk2020a}. While GLCs are line (graph) based \cite{Kovalerchuk2018}, others are pixel-based lossless visualizations \cite{Kovalerchuk2020a} which include GLC-R presented in \cite{Estivill-Castro2020}, and (in this volume) in  \emph{Deep Learning Image Recognition for Non-images} by Kovalerchuk, Kalla, and Agarwal. Also, it will be interesting to see if other lossless visualizations of n-D data can be developed based on entirely different principles.  

\item[\textbf{Problem 2}] How to \emph{select a lossless visualization} of multidimensional data among multiple ones for visual knowledge discovery?  There is no silver bullet, and specific types of lossless visualizations will always serve specific data types.

\item[\textbf{Problem 3}]  \emph{Discovering patterns} in lossless 2-D representations of n-D data. In essence, this is to develop specific machine learning algorithms for such 2-D representations. These 2-D representations are very specific relative to traditional n-D representations—they rather like discovering patterns in pattern recognition on images, including comparing and matching with templates. In general, 2-D spatial representations play an important role in human reasoning, while spatial reasoning is a special case.
\par
While traditional pattern recognition on images deal with the raster images, GLCs form vector images, therefore specific methods are needed for vector images. The area of matching vector images is known as the map conflation area \cite{Kovalerchuk2004}, where a vector road network map from one source is matched with a vector road network map from another source. A similar task is when a vector road network map is matched with the overhead imagery of the area \cite{Kovalerchuk2004}. 
\par
Thus, multiple existing and new techniques in pattern recognition on vector and raster images can be applied to develop methods specific for total 2-D ML. It has already started, as the publications listed above show. 

\item[\textbf{Problem 4}]  Simplification of visible results, predictive models and presenting them to the users. If the model is very complex, its visual representation is often quite multifaceted and can exceed human perception abilities. 

\item[\textbf{Problem 5}] How to evaluate the accuracy of the predictive model produced in the full 2D ML process? Will we use traditional k-fold cross validation, say with k=10, or will we develop and use new alternative methods specific to full 2D ML? Lossless visualization opens an opportunity to introduce new visual evaluation for ML methods \cite{Kovalerchuk2020}. 

\item[\textbf{Problem 6}] Developing full 2D ML methods for \emph{unsupervised} learning. There are multiple specifics in unsupervised learning in this area. We can borrow multiple approaches already developed in visual analytics where a significant portion of work was done for clustering in parallel coordinates. 

\item[\textbf{Problem 7}] Development of full 2D ML for the data with specific characteristics: imbalanced, missed values, very high resolution, and others. The problems with the imbalanced data and data with missed values are well known in machine learning; dealing with them in the 2-D space ML has its own challenges. Dealing with high-resolution data is a particular problem for visual knowledge discovery in full 2-D ML \cite{Kovalerchuk2021}. Consider a numeric attribute with 5 digits in every two values, like  34567 and 34568, which differ only in the last digit. Visual separation of these values can be beyond the resolution of visualization and visual knowledge discovery. Moreover, it creates significant computational challenges for full 2-D ML \cite{Kovalerchuk2021}, which must be addressed.  

\item[\textbf{Problem 8}] How to optimize the full 2D ML process for efficient human perception and interaction? We need to make the process interactive for the domain experts as a self-service. It will allow domain experts to trust the ML predictive model from the very beginning because they built the model themselves. 

\item[\textbf{Problem 9}] How to interpret Deep Learning (DL) models with full 2D ML methodology? The problem of interpreting deep learning models today involves visualization at the latest stage where heatmaps are used to show the salient elements of the model or to show the importance of the attributes with simple bar charts. A much deeper GLC visualization options are available with a full 2D machine learning approach, as we describe below. The main idea of today's interpretation is first building a black box model and then trying to interpret it in domain terms with the help of domain experts. The \emph{fundamental problem} of this approach is that the deep learning models are commonly expressed in very different terms than used in domain knowledge. This domain knowledge is rarely represented with many ``raw'' attributes like pixels or individual words in image and text classification tasks. Often this mismatch makes it practically \emph{impossible} to interpret the deep learning models in terms of existing domain knowledge. Therefore, a fundamentally new approach is needed to resolve challenges.  To generate such fundamentally new solutions, we first need to uncover the \emph{cause of this mismatch}. A significant source of the DL success in getting high accuracy is using so-called ``raw'' features to eliminate feature engineering by domain experts. It dramatically simplifies, speeds up and ``industrialises'' model development. This significant advantage of DL is also a major obstacle for the interpretation of DL models because only features engineered by domain experts bring domain knowledge to the model. How can we reconcile this \emph{deep contradiction} of deep model methodology? Dropping the use of raw features will return us to traditional machine learning and will nullify DL. Can we continue using only raw features and avoid or decrease conceptual mismatch that prevents interpretation?
The fundamentally new methodology is needed to incorporate \emph{the domain knowledge from the very beginning along with raw features}. The advantage of this methodology is in the opportunity to get benefits from both. 

This leads us to the formulation of \textbf{Problems 10-11}.

\item[\textbf{Problem 10}] How to incorporate the \emph{domain knowledge} from the \emph{beginning} along with basic features to discover DL models?  Can total 2D ML help with this?

\item[\textbf{Problem 11}] How to (1) discover new domain knowledge (interpretable features, rules, models) with total 2D ML on raw data and then (2) to use them to \emph{guide} discovering deep learning models on raw data?

Part (1) is \textbf{Problem 3} above with specifics that discovered knowledge must be applied to solve (2). One of the possibilities is starting not from randomly generated weights in discovering DL models but with the higher values of the weights in the areas of raw data where interpretable features are discovered in (1). This will increase the chances that the deep learning models will be interpretable. The goal is to exceed the LIME approach \cite{Ribeiro2016} that builds local linear classifiers in raw data assuming that any linear model is interpretable, which is not the case in general. It is typically applicable for homogeneous data, but data are heterogeneous in machine learning, especially in medical applications (temperature, blood pressure, pulse, weight, height).

\end{description}

\section{Visualization in NLP}
\label{sec:NLP}

A review of visualization techniques for text analysis can be found in \cite{Liu2018}. According to this study, 263 text visualization papers and 4,346 text mining papers were published between 1992 and 2017. The authors derived around 300 concepts (visualization techniques, mining techniques, and analysis tasks) and built taxonomies. The visualization concepts that have the most significant number of related papers are typographic visualizations (text highlighting, word cloud), chart visualizations (bar chart, scatterplot, and graph visualizations (node-link, tree, matrix).

More recently, there has been an increasing interest in visualizing NLP ML models. Using visualization methods in image processing ML tasks is quite intuitive, and saliency maps are a popular tool for gaining insight into the deep learning of images. For instance, Grad-Cam \cite{Selvaraju2017} is a standard technique for saliency map visualizations in deep learning. NLP ML visualization is not intuitive when it is applied to text. 

NLP models are currently visualized by heatmaps similar to Grad-CAM, looking at the connections between tokens in models that utilize attention mechanisms (see \emph{Visualizing and Explaining Language Models} by Bra{\c{s}}oveanu and Andonie, in this volume). These solutions typically look at the importance of individual tokens (words) on the model output. The goal of NLP MLP visualizations is usually to highlight the most significant tokens that have the most important impact on the model output. For instance, what tokens contribute most to the decision that a text is classified as ``fake news'' \cite{Brasoveanu2020}.

The landscape of NLP has recently changed with the introduction of Transformers (see \emph{Visualizing and Explaining Language Models} by Bra{\c{s}}oveanu and Andonie, in this volume). Transformer models can extract complex features from the input data and effectively solve NLP problems. For instance, BERT is a state-of-the-art NLP model developed by Google successfully in NLP tasks such as text classification and sentence prediction.

Explaining the information processing flow and results in a Transformer is difficult because of its complexity. A convenient and very actual approach is visualization. The first survey on visualization techniques for Transformers is \emph{Visualizing and Explaining Language Models} by Bra{\c{s}}oveanu and Andonie (in this book). None of the current visualization systems is capable to examine all the facets of the Transformers, since this area is relatively new and there is no consensus on what needs to be visualized. The visualization of NLP neural models is still under development.

Without being exhaustive, we state the following three open problems in NLP ML visualization:

\begin{description}

\item[\textbf{Problem 1}]  Visualization is often the bridge that language model designers use to explain their work. For instance, coloring of the salient words or sentences, and clustering of neuron activations can be used to quickly understand a neural model. The approach in \emph{Visualizing and Explaining Language Models} by Bra{\c{s}}oveanu and Andonie (in this volume) attempts to visualize Transformer representations of n-tuples, equivalent to context-sensitive text structures. It also showcases the techniques used in some of the most popular deep learning techniques for NLP visualizations, with a special focus on interpretability and explainability. Going beyond current visualizations that are model-agnostic, future frameworks will have to provide visualization components that focus on the important Transformer components like corpora, embeddings, attention heads or additional neural network layers that might be problem-specific. For instance, Yun \emph{et al.} applied dictionary learning techniques to provide detailed visualizations of Transformer representations and insights into the semantic structures \cite{Yun2021}. We consider visualization of context-sensitive syntactic information and semantic structures as one of the hottest applications of ML in NLP. 

\item[\textbf{Problem 2}] There is a subtle interplay between syntactic and semantic information, as outlined in semiotics. In \emph{semiotics}, a \emph{sign} is anything that communicates a meaning, that is not the sign itself to the interpreter of the sign. In-depth definitions can be found in \cite{Morris1972, Bense1975, Eco1976,Sebeok1994,Chandler2017}. The triadic model of semiosis, as stated by Charles Sanders Peirce defined semiosis as an irreducible triadic relation between Sign-Object-Interpretant \cite{Peirce1960}. The recent interest in self-explaining ML models can be regarded as  exposure of the self-interpretation and semiotic awareness mechanism \cite{Musat2020}. The concept of sign and semiotics offers a promising and tempting conceptual basis to ML and visualization.

\item[\textbf{Problem 3}] We have to make a difference between attention and explanation. These terms are frequently used in ML visualization, for instance, when using saliency maps in deep learning visualization. Such saliency maps generally visualize our attention but do not ``explain'' the deep learning model \cite{Jain2019}. A challenging and open problem is to use visualization as an explanation tool for ML models. In other words, visualising the combination of words according to which the text is classified as ``fake news'' highlights our attention or even may explain the classification decision.

\end{description}

\section{Multidimensional Visualizations and Artificial Intelligence}
\label{sec:MVAI}

Visualization plays an essential role in humans' cognitive process~ \cite{ND-7-01}. As a tool, visualization helps us mitigate our limitations regarding information overload~\cite{ND-7-02}; by properly transforming data into information, we will ultimately turn it into new knowledge. However, the increase of the available data, generated at different velocities, and presented in many formats, turns a visual representation of most of the current datasets into a difficult task. At the turn of the century, multidimensional datasets were stored into a special kind of database, known as data warehouses, where data were classified into two buckets: facts and dimensions. Datawarehouse was built to scale and present the users' data in an intelligible way. Using a simple yet powerful visualization solution --- the pivot table and later the pivot chart --- helped many decision-makers get insights into data through Online Analytical Processing (OLAP)~\cite{ND-7-03}. The data are aggregated before being displayed to the analysts. This mechanism allows the pivot table to scale to any kind of dimension, and together with the drill-down (to get more detail) and drill-up to get less detail) it helps decision-makers to explore data. We can see the relation with Schneiderman' visualization mantra. 

Datawarehouse also adopts metadata everywhere used by the visualization interfaces to help users hide data in certain operations and suggest which data goes into the specific part of the pivot table. This usage of contextual (meta) data lowers some manual labor of the users and improves analytical results. Their strength was to have a small concepts model, a single visualization scheme, and a set of restricted interactions. Although datawarehouses and OLAP are still viable tools for some multidimensional data analysis, they assume that everything can be put into a tabular form. Each dimension has a proper semantic and is intelligible to the analysts. 
\begin{figure}[t]
    \centering
    \includegraphics[width=1\textwidth]{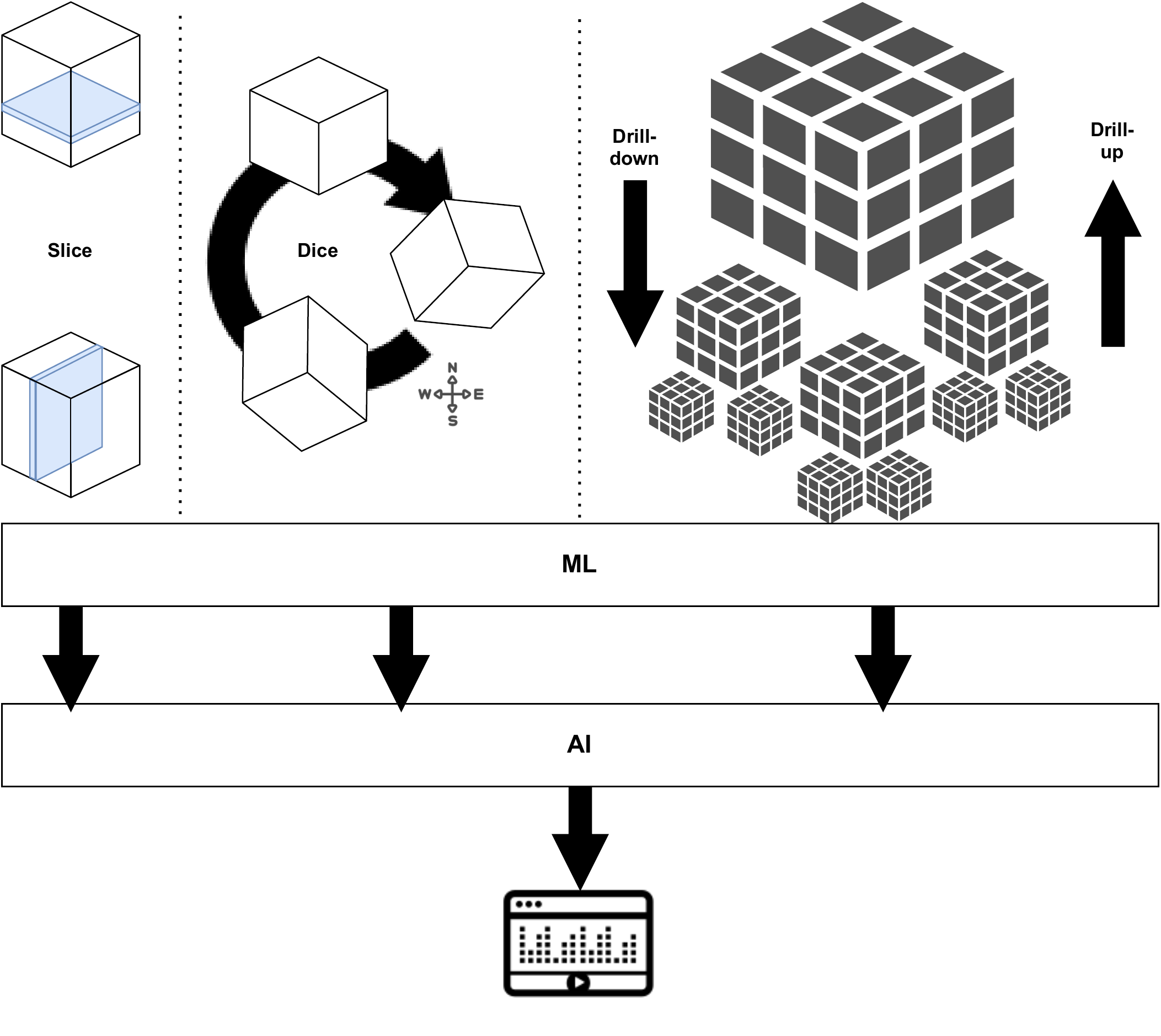}
    \caption{ML and AI role improving the selection of valid data and visualization for multidimensional datasets. While slice, dice and drill up and down are taken by users, ML/AI should do this automatically to settle the center point of the dataset and allow further exploration by the users from there.}
    \label{fig:OLAP-AI-ML}
\end{figure}
This is not the case with many of the datasets we deal with:
\begin{enumerate}
    \item Many datasets result from non-human data harvesting, with high requirements on processing velocity~\cite{ND-7-04}.
\item Data comes in a variety of formats. They are often unstructured, many of which are non-tabular. It is not easy to aggregate such data without knowing the specific goal.
\item Many variables of the datasets result from feature generation algorithms (e.g. medical imaging), whose semantics are not apparent to the human analyst.
\item Multidimensional datasets, especially the high dimensional ones, often are highly sparse.
\end{enumerate}
We need to improve artificial intelligence (AI) solutions to help users with better and fast data analysis towards visual knowledge exploration (See Fig.~\ref{fig:OLAP-AI-ML}). AI is about learning, reasoning, and evolving. 

From the visualization viewpoint, learning (the Machine Learning part of the AI, and often wrongly used as a synonym) is about learning to visualize better, helping the analysts perform the visual exploration. In contrast, from the machine learning viewpoint, visualization is about visualizing machine learning better.  It is important to establish a formal definition of visualization paradigms, namely GLC, to ensure that AI and ML can leverage the data and the formal representation to automatically (or semi-automatically) choose the proper processing steps towards the ``proper'' visualization. 

\begin{figure}[t]
    \centering
    \includegraphics[width=0.65\textwidth]{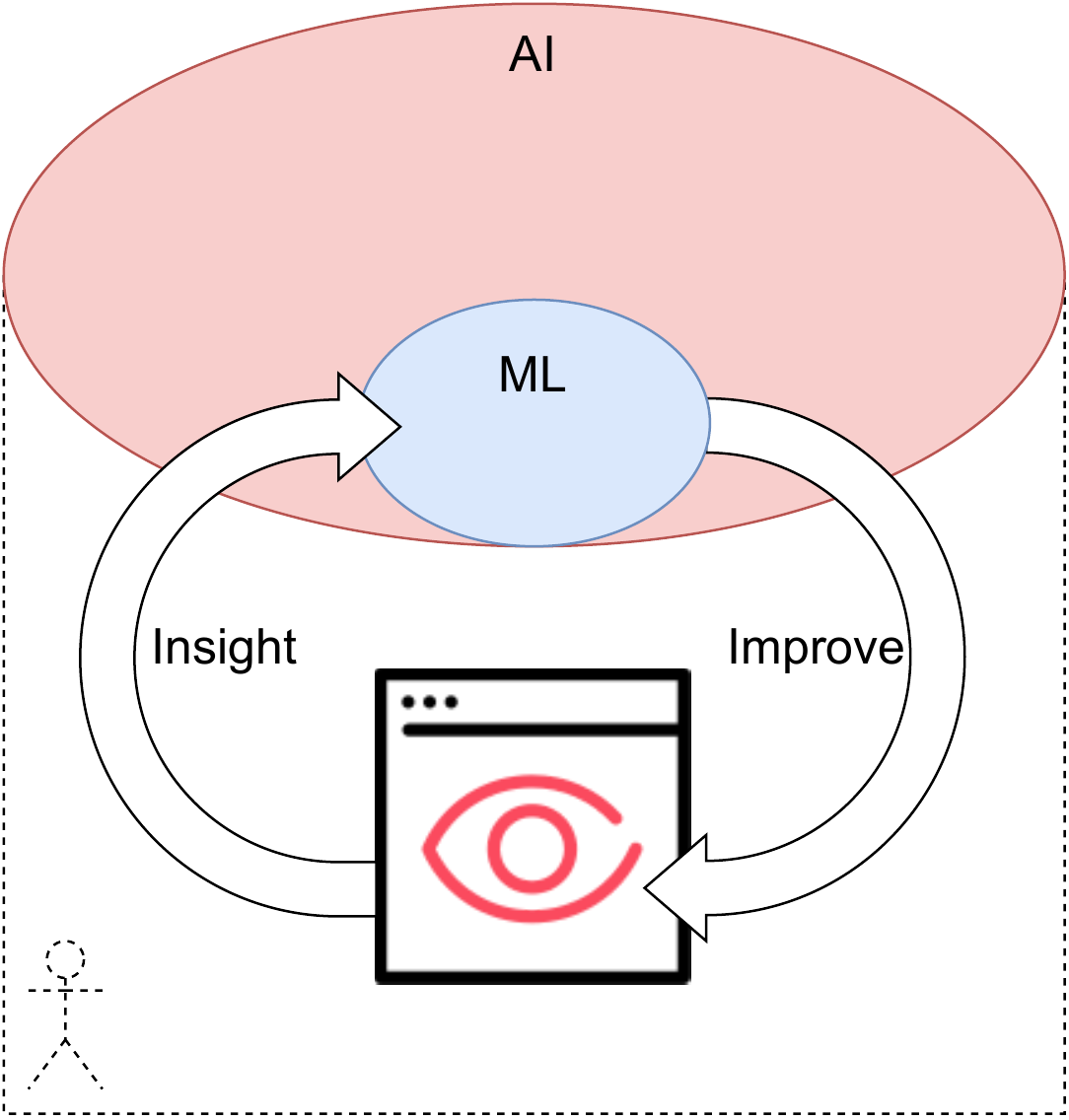}
    \caption{Relation between AI, ML and Visualization towards the Visual Knowledge Extraction}
    \label{fig:AI-ML-VIS}
\end{figure}

The reasoning is the capability to use the known knowledge to derive, in context, solutions or new knowledge. Furthermore, Evolving can increase its awareness of the world by updating the knowledge base, removing non-relevant facts and including new ones. Fig.~\ref{fig:AI-ML-VIS} illustrates the relation between AI, Machine Learning( ML), and Visualization in the Visual Knowledge Exploration carried on by an analyst. 

The context, identified by the dashed line, must be known both to the analyst and the AI algorithms. Thus, contextual data must be stored using an AI tractable representation, such as descriptive logic assertions~\cite{ND-7-05}. That way, every step towards a better visualization can be AI-assisted by using reasoning and optimization, among others. On the one hand, using an interactive approach, humans can still derive new knowledge by interacting in a mixed reality environment, using contextual information to make decisions, assisted by machine learning models. On the other hand, visualization is key to bridge the gap between computer-generated knowledge/models and human knowledge. The computer-generated knowledge can be transferred to humans through our cognitive processes if one can visualize how it is constructed. It is not a question of how good a machine learning model is but instead pursuing the capabilities to say why the model presents such an answer. 

\begin{description}

\item[\textbf{Problem 1}] Level of detail --- Multidimensional datasets present features in multiple levels of detail (LoD). However, discovering new hidden patterns can be difficult depending on the level of detail presented to the user~\cite{ND-7-06}. To analysts, visualization tools need to incorporate AI and ML algorithms that automatically choose, in the background, the proper LoD to start with the interactive analysis. However, to decide which LoD is the best, a goal function must be minimized/maximized. How can we define them, depending on the task and the visualization scheme?

\item[\textbf{Problem 2}] Spatiotemporal patterns --- Datasets representing human activity often contain features related to space and time. Thus, the patterns are spatiotemporal. However, to explain why those patterns occur, we need data about many aspects that do not have a proper and standard representation in time and space. Besides, answers may result from applying an ML model to the data, which sometimes transforms the feature space into another multidimensional space, making representing the results difficult. 

\item[\textbf{Problem 3}] Semantics for visualization --- Most AI reasoning is only applicable if there are proper semantics on the terms used. A knowledge base (KB) is responsible for keeping the information and knowledge used to describe a specific domain's context. It consists of axioms describing conceptual entities and their relations and a set of asserted facts specific to a domain. OLAP has proper semantics and a limited visualization scheme. It can support many analyses, but the lack of visual expressiveness is evident. In order to use complex visualization schemes, there must be a formal definition of the process, a formal definition of the actions, a formal definition of the visual scheme, to name a few. No standard formalization of the visualization aspects exists, making it difficult to have a general solution for visual knowledge exploration. Thus, the task is how to introduce semantics to be able to optimize visualization. 

\item[\textbf{Problem 4}] Transfer of visualization knowledge between domains with AI -  To have AI-supported Visualization in an interactive environment, reducing the scope is key. A set of formal definitions must be defined for a specific domain, together with a set of key visualizations schemes. However, it is not easy to share knowledge between domains when this is done, even if they share many terms and semantics. Thus, a minimum model of interoperability is needed.

\item[\textbf{Problem 5}] Interpretability of discovered patterns - In a Visual Knowledge Discovery environment, even if the analysts discover some patterns of insight, they must be turned into a set of facts that support those findings. For any domain, including social sciences~\cite{ND-7-07} and medicine~\cite{ND-7-08}, the AI-generated result must be interpreted by a human. The interpretability should be done visually to use all our cognitive strengths.

\end{description}
	
\section{Conclusions}
\label{sec:conclusion}

This chapter summarizes the current trend and a view of future directions infusion of AI/ML and visualization. First-time mutual enhancements of AI and Machine Learning (AI/ML) with visualization were codified under \emph{Visual Analytics} over 20 years ago, where AI/ML is considered among other analytical methods. In this volume, we use \emph{Visual Knowledge Discovery} to represent the current fusion stage of AI/ML with visualization, where knowledge discovery with supervised ML plays a critical role. From our viewpoint, the term Visual Knowledge Discovery emphasizes the desired results -- \emph{knowledge discovered}, while Visual Analytics  emphasizes the analytical process.

Recently the prominence of visualization in AI/ML became very evident with the progress in deep learning and machine learning in general, where visualization is a key player in explaining black-box learning models. The following important emerging area of Visual Knowledge Discovery is total ML in the lossless 2D/3D visualization space. It allows building ML algorithms and models, which work on n-D data but in the 2D visualization space without loss of multidimensional information. This emerging capability dramatically expands the opportunities for end-users to build ML models themselves as a self-service bringing deep domain knowledge to the process of explainable ML model discovery.

\bibliographystyle{unsrtnat}
\bibliography{bib.bib}  %%% Uncomment this line and comment out the ``thebibliography'' section below to use the external .bib file (using bibtex) .





% \end{thebibliography}

\end{document}